\documentclass[11pt,a4paper,dvipsnames]{article}
\usepackage{emnlp2021}
\usepackage{times}
\usepackage{latexsym}

\usepackage{microtype}

\usepackage{amsmath}
\usepackage{tikz}
\usetikzlibrary{positioning}
\usepackage{standalone}
\usepackage{booktabs}
\usepackage{graphicx}
\usepackage{multirow}
\usepackage{adjustbox}
\usepackage{algorithm}
\usepackage{amssymb}
\usepackage{bbm}
\usepackage[noend]{algpseudocode}
\usepackage{float}
\newfloat{algorithm}{t}{lop}
\usepackage{multicol}
\newcommand{\pluseq}{\mathrel{+}=}
\DeclareMathOperator*{\softmax}{softmax}
\renewcommand{\Pr}{\mathrm{P}}
\newcommand{\Pins}{\mathrm{P}_\text{ins}}
\newcommand{\Pdel}{\mathrm{P}_\text{del}}
\newcommand{\Psubs}{\mathrm{P}_\text{subs}}

\newcommand{\A}[1]{\ifdim#1pt<10pt~~\fi#1}
\newcommand{\R}[2]{\ifdim#1pt<10pt~\fi#1 \scalebox{0.8}{\tiny $\pm$}{\scriptsize #2}}
\newcommand{\subsarrow}[1][4pt]{\mathrel{%
   \vcenter{\hbox{\rule[-.5\fontdimen8\textfont3]{#1}{\fontdimen8\textfont3}}}%
   \mkern-4mu\hbox{\usefont{U}{lasy}{m}{n}\symbol{41}}}}

\newcommand{\lnwidth}{0.3}

\newcommand{\ins}[1]{\tikz[baseline]{\node[anchor=base,fill=Green!20,inner sep=1.2pt,draw=Green,line width=\lnwidth] {\color{Green}+#1};}}
\newcommand{\del}[1]{\tikz[baseline]{\node[anchor=base,fill=Blue!20,inner sep=1.2pt,draw=Blue,line width=\lnwidth] {\color{Blue}-#1};}}

\newcommand{\subs}[2]{\tikz[baseline]{\node[anchor=base,fill=Dandelion!20,inner sep=1.2pt,draw=Dandelion,line width=\lnwidth] {\textcolor{Blue}{#1}$\subsarrow$\textcolor{Green}{#2}};}}

\newcounter{notecounter}
\newcommand{\enotesoff}{\long\gdef\enote##1##2{}}
\newcommand{\enoteson}{\long\gdef\enote##1##2{{
			\stepcounter{notecounter}
			\large\bf
			\hspace{100cm}\arabic{notecounter} $<<<$ ##1: ##2
			$>>>$\hspace{1cm}}}}
\enoteson
\enotesoff

\usepackage{amssymb}

\usepackage[normalem]{ulem}
\def\OD#1{{\color{cyan!80!yellow!80!black!100}OD: \it #1}}
\def\ODdel#1{\bgroup\markoverwith{\textcolor{cyan!89!yellow!80!black!100}{\rule[0.4ex]{2pt}{3pt}}}\ULon{#1}}
\makeatletter
\def\ODwtf#1{\bgroup \markoverwith{\lower3.5\p@\hbox{\sixly\textcolor{cyan!89!yellow!80!black!100}{\char58}}}\ULon{#1}\OD{??}}
\makeatother

\title{Neural String Edit Distance}

\author{Jindřich Libovický$^1$ \and Alexander Fraser$^2$ \\
$^1$Faculty of Mathematics and Physics, Charles University, Prague, Czech Republic \\
$^2$Center for Information and Language Processing, LMU Munich, Germany \\
  \texttt{libovicky@ufal.mff.cuni.cz}\quad\texttt{fraser@cis.lmu.de}}

\date{}

\begin{document}

\maketitle

\begin{abstract}
We propose the \emph{neural string edit distance} model for string-pair
    matching and string transduction based on learnable string edit distance.
We modify the original expectation-maximization learned edit distance algorithm
    into a differentiable loss function, allowing us to integrate it into a
    neural network providing a contextual representation of the input.
We evaluate on cognate detection, transliteration, and grapheme-to-phoneme
    conversion, and show that we can trade off between performance and
    interpretability in a single framework.
Using contextual representations, which are difficult to interpret, we match
    the performance of state-of-the-art string-pair matching models.
Using static embeddings and a slightly different loss function, we force
    interpretability, at the expense of an accuracy drop.
\end{abstract}

% ============================================================================
\section{Introduction}
% ============================================================================

%\enote{AF}{If you need space, you can move the information in the footnotes to the appendix.}

%\enote{AF}{I rewrote the end of the abstract to say that if we use black-box, we get good results, and if we use context-free, we get good interpretability.
%  If this looks good, I can update the sections at the beginning of the paper with this message. JL: Yes, I like it.}
  %
%\enote{AF}{Possibly the writing in the abstract can be strengthened further to argue that it is good to have both of these possibilities in the same framework (but I am not sure we should push this argument strongly, what do you think?). JL: sounds unnecessary to me}

State-of-the-art models for string-pair classification and string transduction
employ powerful neural architectures that lack interpretability.
%E.g.,
For example,
BERT \citep{devlin-etal-2019-bert}
%internally
compares all input symbols with
each other via 96 attention heads, whose functions are difficult to interpret.
Moreover, attention itself can be hard to interpret
\citep{jain-wallace-2019-attention,wiegreffe-pinter-2019-attention}.

In many tasks, such as in transliteration, a relation between two strings can
be interpreted more simply as edit operations \citep{levenshtein1966binary}.
The edit operations define the alignment between the strings and provide an
interpretation of how one string is transcribed into another. Learnable edit
distance \citep{ristad1998learning} allows learning the weights of edit
operations from data using the expectation-maximization (EM) algorithm. Unlike
post-hoc analysis of black-box models,
%that
which depends on human qualitative judgment
\citep{adadi2018peeking,hoover-etal-2020-exbert,lipton2018mythos}, the
restricted set of edit operations allows direct interpretation.
%of the model operation.
%
Unlike hard attention \citep{mnih2014recurrent,indurthi-etal-2019-look} which
also provides a discrete alignment between input and output,
%the
edit distance explicitly says how the input symbols are processed.
Also, unlike models like Levenshtein Transformer \citep{gu2019levenshtein},
which does not explicitly align source and target uses edit operations to model
intermediate generation steps only within the target string, learnable edit
distance considers both source and target symbols to be a subject of the edit
operations.

% % % % % % % % % % % % % % % % % % % % % % % % % % % % % % % % % % % % % % %
\begin{figure}
    \centering
    %\resizebox{0.95\columnwidth}{!}{\trimbox{0.7cm 0.2cm 0.2cm 0.2cm}{\includestandalone{scheme}}}
    \includegraphics[width=.95\columnwidth]{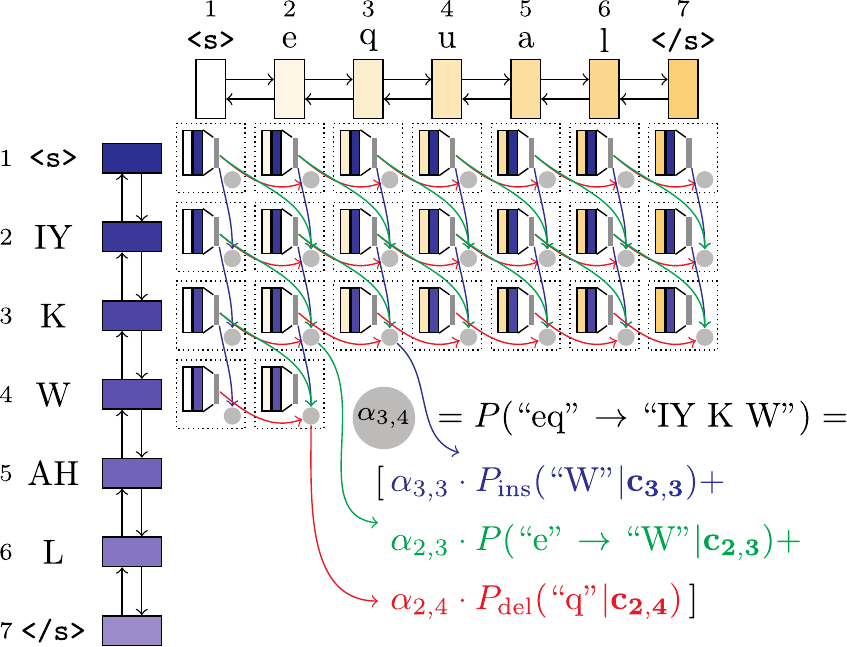}

    \caption{
      An example of applying
      the dynamic programming algorithm used to compute the
    edit probability score. It gradually fills the table of probabilities that
    prefixes of the word ``equal'' transcribe into prefixes of phoneme sequence
    ``IY K W AH L''. The probability (gray circles) depends on the
    probabilities of the prefixes and probabilities of plausible edit
    operations: insert (blue arrows), substitute (green arrows) and delete (red
    arrows).}\label{fig:scheme}
\end{figure}
% % % % % % % % % % % % % % % % % % % % % % % % % % % % % % % % % % % % % % %

We reformulate the EM training used to train learnable edit distance as a
differentiable loss function that can be used in a neural network. We propose
two variants of models based on \emph{neural string edit distance}:
%first,
a
bidirectional model for string-pair matching
%, second,
and
a conditional model for
string transduction.
%, and
We
evaluate
%these on the tasks of
on
cognate detection,
transliteration, and grapheme-to-phoneme (G2P) conversion. The model jointly
learns to perform the task and to generate a latent sequence of edit operations
explaining the output. Our approach can flexibly trade off performance and
intepretability by using input representations with various degrees of
contextualization and outperforms methods that offer a similar degree of
interpretability \citep{tam-etal-2019-optimal}.

% ============================================================================
\section{Learnable Edit Distance}
% ============================================================================

Edit distance \citep{levenshtein1966binary}
%
%generalized for two different alphabets
%
formalizes transcription of a string $\mathbf{s} = (s_1, \ldots,
s_n)$ of $n$ symbols from alphabet $\mathcal{S}$ into a string $\mathbf{t} =
(t_1, \ldots, t_m)$ of $m$ symbols from alphabet $\mathcal{T}$ as a sequence of
operations: delete, insert and substitute, which
%can
have different costs.

\citet{ristad1998learning} reformulated operations as random
events drawn from a distribution of all possible operations: deleting any $s
\in \mathcal{S}$, inserting any $t \in \mathcal{T}$, and substituting any pair
of symbols from $\mathcal{S} \times \mathcal{T}$. The probability
$\Pr(\mathbf{s}, \mathbf{t}) = \alpha_{n,m}$ of $\mathbf{t}$ being
%the result of editing of
edited from
$\mathbf{s}$
%and
can be expressed recursively:
\begin{eqnarray}
    \label{eq:alpha} \alpha_{n,m} & = & \alpha_{n,m-1} \cdot \Pins(t_m) + \\
    \nonumber & & \alpha_{n-1, m} \cdot \Pdel(s_n) + \\
    %
%    \nonumber & & \alpha_{n-1, m-1} \cdot \Psubs(s_{n-1}, t_m)
   \nonumber & & \alpha_{n-1, m-1} \cdot \Psubs(s_n, t_m)
\end{eqnarray}

This can be computed using
the
dynamic programming
algorithm
of \citet{wagner1974string},
which also computes values of $\alpha_{i,j}$ for all prefixes
$\mathbf{s}_{:i}$ and $\mathbf{t}_{:j}$.
%Because t
The operation probabilities
only depend on the individual pairs of symbols at positions $i$, $j$,
so
the same
dynamic programming algorithm
%can be
is
used for computing the \emph{suffix-pair}
transcription probabilities $\beta_{i,j}$ (the backward probabilities).
%
% TODO in cameara ready, add the beta algorithm into Appendix

With a training corpus of pairs of matching strings, the operation
probabilities can be estimated using
the
EM algorithm.  In the expectation step,
expected counts of all edit operations are estimated for the current parameters
using the training data. Each pair of symbols $s_i$ and $t_j$ contribute to the
expected counts of the operations:
\begin{equation}
    E_\text{subs}(s_i, t_j) \pluseq \alpha_{i-1,j-1} \Psubs(s_i,
    t_j)\beta_{i,j} / \alpha_{n,m}\label{eq:statExpectation}
\end{equation}
and analogically for the delete and insert operations. In the maximization
step,
%the
operation probabilities are estimated by normalizing the expected
counts.
%For more details, we refer the reader to
See
Algorithms 1--5 in
\citet{ristad1998learning}
%.
for more details.

% ============================================================================
\section{Neural String Edit Distance Model}
% ============================================================================

In our model, we replace the discrete table of operation probabilities with a
probability estimation based on a continuous representation of the input, which
brings in the challenge of changing the EM training into a differentiable loss
function that can be back-propagated into the representation.

Computation of the transcription probability is shown in Figure~\ref{fig:scheme}.
We use the same dynamic programming algorithm (Equation~\ref{eq:alpha} and
Algorithm~\ref{alg:alpha} in Appendix~\ref{ap:alg}) that gradually fills a
table of probabilities row by row. The input symbols are represented by
learned, possibly contextual embeddings (yellow and blue boxes in
Figure~\ref{fig:scheme}) which are used to compute a representation of symbol
pairs with a small feed-forward network. The symbol pair representation is used
to estimate the probabilities of insert, delete and substitute operations
(blue, red and green arrows in Figure~\ref{fig:scheme}).

Formally, we embed the source sequence $\mathbf{s}$ of length $n$ into a matrix
$\mathbf{h^s} \in \mathbb{R}^{n\times d}$ and analogically $\mathbf{t}$ into
$\mathbf{h^t} \in \mathbb{R}^{m\times d}$ (yellow and blue boxes in
Figure~\ref{fig:scheme}). We represent the symbol-pair contexts as a function of
the respective symbol representations (small gray rectangles in
Figure~\ref{fig:scheme}) as a function of repspective symbol representation
$\mathbf{c}_{i,j} = f(\mathbf{h}^s_i, \mathbf{h}^t_j)$ depending on the task.

The logits (i.e., the probability scores before normalization) for the edit
operations are obtained by concatenation of the following vectors (corresponds
to red, green and blue arrows in Figure~\ref{fig:scheme}):
\begin{itemize}

    \setlength{\itemsep}{2pt}

\item $\mathbf{z}_\text{del}^{i,j} = \text{Linear}(\mathbf{c}_{i-1,j}) \in \mathbb{R}^{d_\text{del}}$,

\item $\mathbf{z}_\text{ins}^{i,j} = \text{Linear}(\mathbf{c}_{i,j-1}) \in \mathbb{R}^{d_\text{ins}}$,

\item $\mathbf{z}_\text{subs}^{i,j} = \text{Linear}(\mathbf{c}_{i-1,j-1}) \in \mathbb{R}^{d_\text{subs}}$,

\end{itemize}
where $\text{Linear}(\mathbf{x}) = \mathbf{W}\mathbf{x} + \mathbf{b}$ where
$\mathbf{W}$ and $\mathbf{b}$ are trainable parameters of a linear projection
and $d_\text{del}$, $d_\text{ins}$ and $d_\text{subs}$ are the numbers of
possible delete, insert and substitute operations given the vocabularies.
The distribution $\Pr_{i,j} \in \mathbb{R}^{d_\text{del} + d_\text{ins} +
d_\text{subs}}$ over operations that lead to prefix pair $\mathbf{s}_{:i}$ and
$\mathbf{t}_{:j}$ in a single derivation step is
\begin{equation}
    \Pr_{i,j} = \softmax(\mathbf{z}_\text{del}^{i,j} \oplus
    \mathbf{z}_\text{ins}^{i,j} \oplus \mathbf{z}_\text{subs}^{i,j}).\label{eq:distr}
    i,j%
\end{equation}
The probabilities $\Pdel^{i,j}$, $\Pins^{i,j}$ and $\Psubs^{i,j}$ are obtained
by taking the respective values from the distribution corresponding to the
logits.\footnote{Using Python-like notation $\Pdel^{i,j} =
\Pr_{i,j}\mathtt{[}:d_\text{del}\mathtt{]}$, \\\phantom{xxxxx}$\Pins^{i,j} =
\Pr_{i,j}\mathtt{[}d_\text{del}:d_\text{del}+d_\text{ins}\mathtt{]}$,
$\Psubs^{i,j} = \Pr_{i,j}\mathtt{[}d_\text{del}+d_\text{ins}:\mathtt{]}$.}
Note that $\Pr_{i,j}$ only depends on (possibly contextual) input embeddings
$\mathbf{h}^s_i$, $\mathbf{h}^s_{i-1}$, $\mathbf{h}^t_j$, and
$\mathbf{h}^t_{j-1}$, but not on the derivation of prefix $\mathbf{t}_{:j}$
from $\mathbf{s}_{:i}$.

The transduction probability $\alpha_{i,j}$, i.e., a probability that
$\mathbf{s}_{:i}$ transcribes to $\mathbf{t}_{:j}$ (gray circles in
Figure~\ref{fig:scheme}) is computed in the same way as in
Equation~\ref{eq:alpha}.

The same algorithm with the reversed order of iteration can be used to compute
probabilities $\beta_{i,j}$, the probability that suffix $\mathbf{s}_{i:}$
transcribes to $\mathbf{t}_{j:}$. The complete transduction probability is the
same, i.e., $\beta_{1,1} = \alpha_{n,m}$. Tables $\mathbf{\alpha}$ and
$\mathbf{\beta}$ are used to compute the EM training loss
$\mathcal{L}_\text{EM}$ (Algorithm~\ref{alg:em}) which is then optimized using
gradient-based optimization. Symbol~$\bullet$ in the probability stands for all
possible operations (the operations that the model can assign a probability
score to), ``normalize''' means scale the values such that
they
sum up to one.

% % % % % % % % % % % % % % % % % % % % % % % % % % % % % % % % % % % % % % %
\begin{algorithm}
\caption{Expectation-Maximization Loss}\label{alg:em}
\small
\begin{algorithmic}[1]
	\State $\mathcal{L}_{\text{EM}}$ $\gets 0$
    \For{$i = 1 \ldots n$}
        \For{$j = 1 \ldots m$}
            \State plausible $\gets \mathbf{0}$
            \Comment{Indication vector} \\ \Comment{I.e., operations that can be used given $s_i$ and $t_j$}

            \If{$j > 1$} \Comment{Insertion is plausible}
                \State plausible $\pluseq \mathbbm{1}(\text{insert~} t_j)$
                \State $E^\text{ins}_{i,j} \gets \alpha_{i, j-1} \cdot \Pins(\bullet |\mathbf{c}_{i,j-1}) \cdot \beta_{i,j}$
            \EndIf
            \If{$i > 1$} \Comment{Deletion is plausible}
                \State plausible $\pluseq \mathbbm{1}(\text{delete~} s_i)$
                \State $E^\text{del}_{i,j} \gets \alpha_{i-1, j} \Pdel(\bullet|\mathbf{c}_{i-1,j}) \beta_{i,j}$
            \EndIf
            \If{$i > 1$ and $j > 1$} \Comment{Subs. is plausible}
                \State plausible $\pluseq \mathbbm{1}(\text{substitute~} s_i \rightarrow t_j)$
                \State $E^\text{subs}_{i,j} \gets \alpha_{i-1,j-1} \cdot \Psubs(\bullet | \mathbf{c}_{i-1,j-1}) \cdot \beta_{i,j}$
            \EndIf
            \State expected $\gets \text{normalize}( \text{plausible~} \odot$ \\
            \hfill $ \left[E^\text{ins}_{i,j} \oplus E^\text{del}_{i,j} \oplus E^\text{subs}_{i,j}\right])$
            \\ \Comment{Expected distr.\ can only contain plausible ops.}
            \State $\mathcal{L}_{\text{EM}}$ $\pluseq \text{KL}(\Pr_{i,j} || \text{~expected})$
        \EndFor
    \EndFor
    \State \Return $\mathcal{L}_{\text{EM}}$
\end{algorithmic}
\end{algorithm}
% % % % % % % % % % % % % % % % % % % % % % % % % % % % % % % % % % % % % % %

Unlike the statistical model that uses a single discrete multinomial
distribution and stores the probabilities in a table, in our neural model the
operation probabilities are conditioned on continuous vectors. For each
operation type, we compute the expected
%operation
distribution given the
$\mathbf{\alpha}$ and $\mathbf{\beta}$ tables (line~6--14). From this
distribution, we only select operations that are plausible given the context
(line~15), i.e., we zero out the probability of all operations that do not
involve symbols $s_i$ and $t_j$.  Finally (line~18), we measure the KL divergence
of the predicted operation distribution $\Pr_{i,j}$ (Equation~\ref{eq:distr}) from
the expected distribution, which is the loss function $\mathcal{L}_\text{EM}$.

With a trained model, we can estimate the probability of $\mathbf{t}$ being a
good transcription of $\mathbf{s}$. Also, by replacing the summation in
Equation~\ref{eq:alpha} by the $\max$ operation, we can obtain the most probable
operation sequence of operation transcribing $\mathbf{s}$ to $\mathbf{t}$ using
the
\citet{viterbi1967error}
%Viterbi
algorithm.
%
% TODO for camera ready, Viterbi pseudo-code into the Appendix

Note that the interpretability of our model depends on how contextualized the
input representations $\mathbf{h}^s$ and $\mathbf{h}^t$ are. The degree of
contextualization spans from static symbol embeddings with the same strong
interpretability as
statistical models, to Transformers with richly
contextualized representations, which, however, makes our model more similar to
standard black-box models.

%\enote{AF}{This is another place where you need to consistently state what
%sort of tradeoff we have (this might already be OK, I am not sure)}

% ----------------------------------------------------------------------------
\subsection{String-Pair Matching}
% ----------------------------------------------------------------------------

Here, our goal is to train a binary classifier deciding if strings $\mathbf{t}$
and $\mathbf{s}$ match. We consider strings matching if $\mathbf{t}$ can be
obtained by editing $\mathbf{s}$, with the probability $\Pr(\mathbf{s},
\mathbf{t}) = \alpha_{n,m}$ higher than a threshold. The model needs to learn
to assign a high probability to derivations of matching the source string to
the target string and low probability to derivations matching different target
strings.

The symbol-pair context $\mathbf{c}_{i,j}$ is computed as
\begin{equation}
    \operatorname{LN}\left(\operatorname{ReLU}\left(\operatorname{Linear}(\mathbf{h}^s_i
    \oplus \mathbf{h}^t_j) \right) \right) \in \mathbb{R}^d,
\end{equation}
where $\operatorname{LN}$ stands for layer normalization and $\oplus$ means
concatenation.

The statistical model assumes a single multinomial table over edit
operations. A non-matching string pair gets little probability because all
derivations (i.e., sequence of edit operations) of non-matching string pairs
consist of low-probability operations and high probability is assigned to
operations that are not plausible. In the neural model, the same information
can be kept in model parameters and we can thus simplify the output space of
the model (see Appendix~\ref{ap:motivation} for thought experiments justifying
the design choices).

We no longer need to explicitly model the probability of implausible operations
and can only use \emph{a single class} for each type of edit operation (insert,
delete, substitute) and one additional \emph{non-match} option that stands for
the case when the inputs strings do not match and none of the plausible edit
operations is probable (corresponding to the sum of probabilities of the
implausible operations in the statistical model).

The value of $\Pr(\mathbf{s}, \mathbf{t}) = \alpha_{m,n}$ serves as a
classification threshold for the binary classification. As additional training
signal, we also explicitly optimize the probability using the binary
cross-entropy as an auxiliary loss, pushing the value towards 1 for positive
examples and towards 0 for negative examples. We set the classification
threshold dynamically to maximize the validation $F_1$-score.

% ----------------------------------------------------------------------------
\subsection{String Transduction}
% ----------------------------------------------------------------------------

In the second use case, we use neural string edit distance as a string
transduction model: given a source string, edit operations are applied to
generate a target string. Unlike classification, we model the transcription
process with vocabulary-specific-operations, but still use only a single class
for deletion. For the insertion and substitution operation, we use
$|\mathcal{T}|$ classes corresponding to the target string alphabet. Unlike
classification, we do not add the non-match class. To better contextualize the
generation, we add attention to the symbol-pair representation
$\mathbf{c}_{i,j}$:
\begin{equation}
    \operatorname{LN}\left(\operatorname{ReLU}\left(\operatorname{Linear}(\mathbf{h}^s_i
    \oplus \mathbf{h}^t_j) \right) \oplus \operatorname{Att}\left(
    \mathbf{h}^t_j, \mathbf{h}^s \right) \right)% \in \mathbb{R}^{2d},
\end{equation}
of dimension $2d$, where $\operatorname{Att}(\mathbf{q}, \mathbf{v})$ is a
multihead attention with queries $\mathbf{q}$ and keys and values $\mathbf{v}$.

While generating the string left-to-right, the only way a symbol can be
generated is either by inserting it or by substituting a source symbol.
Therefore, we estimate the probability of inserting symbol $t_{j+1}$ given a
target prefix $\mathbf{t}_{:j}$ from the probabilities of inserting a symbol
after $t_{j}$ or substituting any $s_i$ by $t_{j+1}$ (i.e., averaging over a
row in Figure~\ref{fig:scheme}):
%j
\begin{gather}
    \label{eq:next} P(t_{j+1} | \hat{\mathbf{t}}_{:j}, \mathbf{s}) =
    \sum_{j=1}^{|S|} \alpha_{i,j} \Pins(t_{j+1} | \mathbf{c}_{i,j}) \nonumber
    \\
    + \sum_{j=2}^{|S|} \alpha_{i,j} \Psubs(s_i, t_{j+1} | \mathbf{c}_{i,j}).
\end{gather}
Probabilities $\Pins$ and $\Psubs$ are respective parts of the distribution
$\Pr_{i,j}$ (Equation~\ref{eq:distr}). Probablity $\Pdel$ is unkown at this point
because computing it would be computed based on state $\mathbf{c}_{i,j+1}$
which is impossible without what the $(j+1)$-th target symbol is, where logits
for $\Pins$ and $\Psubs$ use $\mathbf{c}_{i,j}$ and $\mathbf{c}_{i-1,j}$.
Therefore, we approximate Equation~\ref{eq:distr} as
\begin{equation}
    \hat{\Pr}_{i,j} = \softmax\left( \mathbf{z}_\text{ins}^{i,j} \oplus
    \mathbf{z}_\text{subs}^{i,j} \right).
\end{equation}

At inference time, we decide the next symbol $\hat{t}_j$ based on
$\hat{P}_{i,j}$. Knowing the symbol allows computing the $\Pr_{i,j}$
distribution and values $\alpha_{\bullet,j}$ that are used in the next step of
inference. The inference can be done using the beam search algorithm as is
done with sequence-to-sequence (S2S) models.

We also use the probability distribution $\hat{P}$ to define an additional
training objective which is the \emph{negative log-likelihood} of the ground
truth output with respect to this distribution, analogically to training
S2S models,
\begin{equation}
    \mathcal{L}_\text{NLL} = -\sum_{j=0}^{|\mathbf{t}|} \log
    \sum_{i=0}^{|\mathbf{s}|} \hat{\Pr}_{i,j} / |{\mathbf{s}}|.
\end{equation}

% ----------------------------------------------------------------------------
\subsection{Interpretability Loss}
% ----------------------------------------------------------------------------

In our preliminary experiments with Viterbi decoding, we noticed that the model
tends to avoid the substitute operation and chose an order of insert and delete
operations that is not interpretable. To prevent this behavior, we introduce an
additional regularization loss. To decrease the values of $\mathbf{\alpha}$
that are further from the diagonal, we add the term
$\sum_{i=1}^n\sum_{j=1}^m |i-j| \cdot \alpha_{i,j}$
to the loss function. Note that this formulation assumes that the source and
target sequence have similar lengths. For tasks where the sequence lengths vary
significantly, we would need to consider the sequence length in the loss
function.

In the string transduction model, optimization of this term can lead to a
degenerate solution by flattening all distributions and thus lowering all
values in table $\mathbf{\alpha}$. We thus compensate for this loss by adding
the $-\log\alpha_{n,m}$ term to the loss function which enforces increasing the
$\mathbf{\alpha}$ values.

% % % % % % % % % % % % % % % % % % % % % % % % % % % % % % % % % % % % % % %
\begin{table*}
    %\resizebox{\columnwidth}{!}{\input{cognates_test_table}}
    \centering\scalebox{1.00}{\small
\begin{tabular}{lllccc ccc}
    \toprule
    \multicolumn{2}{l}{\multirow{3}{*}{Method}} &
    \multirow{3}{*}{\# Param.} &
        \multicolumn{3}{c}{Indo-European} &
        \multicolumn{3}{c}{Austro-Asiatic} \\
        \cmidrule(lr){4-6} \cmidrule(lr){7-9}

    \multicolumn{2}{l}{}       &  &
        Plain & + Int.\ loss & Time &
        Plain & + Int.\ loss & Time \\ \midrule

    \multicolumn{2}{l}{Learnable edit distance} & 0.2M &
        \R{32.8}{1.8} & --- & 0.4h  &
        \R{10.3}{0.5} & --- & 0.2h  \\

    \multicolumn{2}{l}{Transformer \tt[CLS]} & 2.7M &
        \R{93.5}{2.1} & --- & 0.7h  &
        \R{78.5}{0.8} & --- & 0.6h  \\ \midrule

    \multirow{3}{*}{\rotatebox{90}{\scriptsize STANCE~~}}
    & unigram     &  0.5M &
        \R{46.2}{4.9} & --- & 0.2h &
        \R{16.6}{0.3} & --- & 0.1h \\ \cmidrule{2-9}
    & RNN         & 1.9M  &
        \R{80.6}{1.2} & --- & 0.3h &
        \R{15.9}{0.2} & --- & 0.2h \\
    & Transformer & 2.7M &
        \R{76.7}{1.3} & --- & 0.3h &
        \R{16.7}{0.3} & --- & 0.2h \\ \midrule

    \multirow{4}{*}{\rotatebox{90}{ours~~}}
    & unigram     &  0.5M &
        \R{78.5}{1.0} & \R{80.1}{0.8} & 1.5h &
        \R{47.8}{0.7} & \R{48.4}{0.6} & 0.7h \\

    & CNN (3-gram) & 0.7M &
        \R{94.0}{0.7} & \R{93.9}{0.8} & 0.9h &
        \R{77.9}{1.5} & \R{76.2}{1.9} & 0.5h
    \\ \cmidrule{2-9}

    & RNN         & 1.9M  &
        \R{96.9}{0.6} & \bf\R{97.1}{0.6} & 1.9h &
        \bf\R{84.0}{0.4} & \R{83.7}{0.5} & 1.2h \\
    & Transformer & 2.7M &
        \R{87.2}{1.6} & \R{87.3}{1.8} & 1.6h &
        \R{69.9}{1.0} & \R{70.7}{1.1} & 1.0h \\
    \bottomrule
\end{tabular}
}

    \caption{F$_1$ and training time
      %scores
      for cognate detection.
      %and respective training times.
      %The F$_1$-scores on the
      F$_1$ on
      validation
      %data are in
      is in
      Table~\ref{tab:cognates_valid} in
      the
      Appendix.}\label{tab:cognates}

\end{table*}
% % % % % % % % % % % % % % % % % % % % % % % % % % % % % % % % % % % % % % %

% ============================================================================
\section{Experiments}
% ============================================================================

We evaluate the string-pair matching model on cognate detection, and the
string transduction model on Arabic-to-English transliteration and English
grapheme-to-phoneme conversion.

In all tasks, we study four ways of representing the input symbols with
different degrees of contextualization.  The interpretable context-free
(unigram) encoder uses symbol embeddings summed with learned position
embeddings. We use a 1-D convolutional neural network (CNN) for locally
contexualized representation where hidden states correspond to consecutive
input $n$-grams. We use bidirectional recurrent networks (RNNs) and
Transformers \citep{vaswani2017attention} for fully contextualized input
representations.

Architectural details and hyperparameters are listed in
Appendix~\ref{ap:hyper}. All hyperparameters are set manually based on
preliminary experiments. Further hyperparameter tuning can likely lead to
better accuracy of both baselines and our model. However, preliminary
experiments showed that increasing the model size only has a small effect on
model accuracy. We run every experiment 5 times and report the mean performance
and the standard deviation to control for training stability.
The source code for the experiments is available at
\url{https://github.com/jlibovicky/neural-string-edit-distance}.

% ----------------------------------------------------------------------------
%\subsection{Cognate Detection}
\paragraph{Cognate Detection.}
% ----------------------------------------------------------------------------
%
Cognate detection is the task of detecting if words in different languages have
the same origin.
We experiment with Austro-Asiatic languages \citep{sidwell2015austroasiatic}
and Indo-European languages \citep{dunn2012indo} normalized into the
international phonetic alphabet as provided by
\citet{rama-etal-2018-automatic}.\footnote{\url{https://www.aclweb.org/anthology/attachments/N18-2063.Datasets.zip}}

For Indo-European languages, we have 9,855 words (after excluding
singleton-class words) from 43 languages forming 2,158 cognate classes. For
Austro-Asiatic languages, the dataset contains 11,828 words of 59 languages,
forming only 98 cognate classes without singletons.
We generate classification pairs from these datasets by randomly sampling 10
negative examples for each true cognate pair. We use 20k pairs for validation
and testing, leaving 1.5M training examples for Indo-European and 80M for
Austro-Asiatic languages.

Many cognate detection methods are unsupervised and are evaluated by comparison
of a clustering
%obtained
from the method with true cognate classes.
%Because w
We
train a supervised classifier,
%we evaluate our method using
so we use
F$_1$-score on our
splits of the dataset.

Because the input and the output are from the same alphabet, we share the
parameters of the encoders of the source and target sequences.

As a baseline we use the original statistical learnable edit distance
\citep{ristad1998learning}. The well-performing black-box model used as another
baseline for comparison with our model is a Transformer processing a
concatenation of the two input strings. Similar to BERT
\citep{devlin-etal-2019-bert}, we use the representation of the first technical
symbol as an input to a linear classifier.
%
%Further, we
We also
compare our results with the STANCE model
\citep{tam-etal-2019-optimal}, a neural model utilizing optimal-transport-based
alignment over input text representation which makes similar claims about
interpretability as we do. Similar to our model, we experiment with various
degrees of representation contextualization.

% ----------------------------------------------------------------------------
%\subsection{Transliteration and G2P Conversion}
\paragraph{Transliteration and G2P Conversion.}
% ----------------------------------------------------------------------------
%
For string transduction, we test our model on two tasks: Arabic-to-English
transliteration
\citep{rosca2016sequence}\footnote{\url{https://github.com/google/transliteration}}
and English G2P conversion using the CMUDict dataset
\citep{weide2005carnegie}\footnote{\url{https://github.com/microsoft/CNTK/tree/master/Examples/SequenceToSequence/CMUDict/Data}}.

The Arabic-to-English transliteration dataset consists of 12,877 pairs for
training, 1,431 for validation, and 1,590 for testing. The source-side alphabet
uses 47 different symbols; the target side uses 39. The CMUDict dataset
contains 108,952 training, 5,447 validation, and 12,855 test examples, 10,999 unique. The
dataset uses 27 different graphemes and 39 phonemes.

We evaluate the output strings using Character Error Rate (CER): the
%average
standard edit distance between the generated hypotheses and the ground truth
string
%, relative to
divided by
the ground-truth string length; and Word Error Rate (WER):
the proportion of words that were transcribed incorrectly. The CMUDict dataset
contains multiple transcriptions for some words,
%following previous work
%, in such cases,
as is usually done
we select the transcription with the lowest CER as a reference.

Unlike the string-matching task, the future target symbols are unknown.
Therefore, when using the contextual representations, we encode the target
string using a single-direction RNN and using a masked Transformer,
respectively.

To evaluate our model under low-resource conditions, we conduct two sets of
additional experiments with the transliteration of Arabic. We compare our
unigram and RNN-based models with the RNN-based S2S model trained on smaller
subsets of training data (6k, 3k, 1.5k, 750, 360, 180, and 60 training
examples) and different embedding and hidden state size (8, 16, \ldots, 512).

For the G2P task, where the source and target symbols can be approximately
aligned, we further quantitatively assess the model's interpretability by
measuring how well it captures alignment between the source and target string.
We consider the substitutions in the Viterbi decoding to be aligned symbols. We
compare this alignment with
%unsupervised
statistical
word
alignment
%that we use as ground truth
and report the F$_1$ score.
We obtain the source-target strings alignment using Efmaral
\citep{ostling2016efmaral}, a state-of-the-art word aligner, by running the
aligner on the entire CMUDict dataset. We use grow-diagonal
%heuristics
for
alignment symmetrization.

The baseline models are RNN-based \citep{bahdanau2015neural} and
Transformer-based \citep{vaswani2017attention} S2S models.

% ============================================================================
\section{Results}
% ============================================================================

% % % % % % % % % % % % % % % % % % % % % % % % % % % % % % % % % % % % % % %
\begin{figure}

    %\resizebox{\columnwidth}{!}{

    \centering
    \scalebox{0.40}{%
        \includegraphics{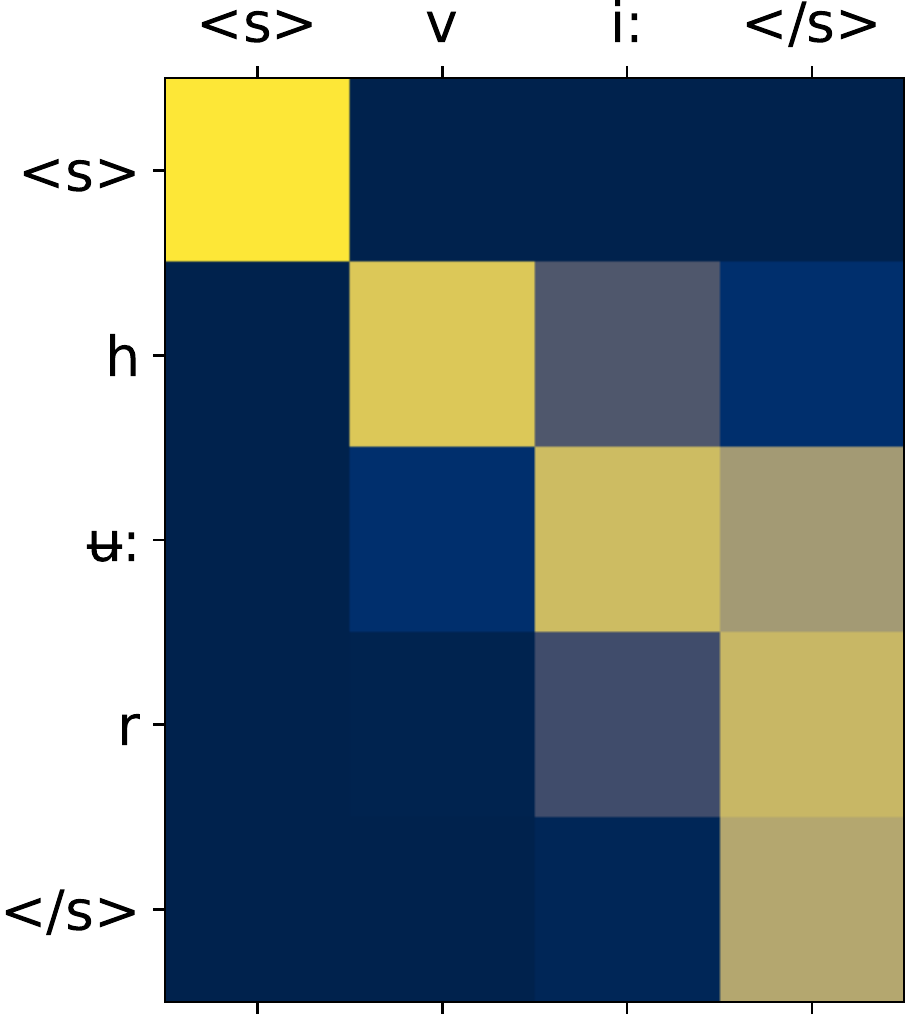}~\includegraphics{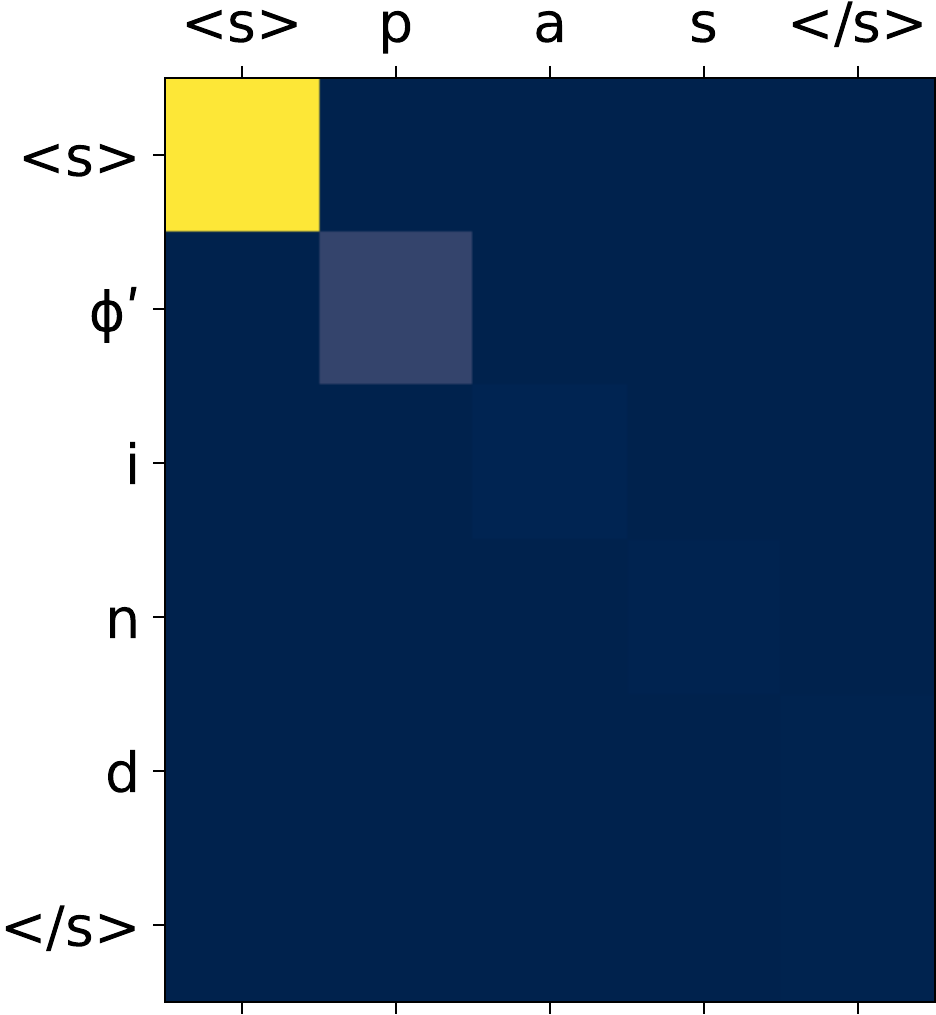}}

    \caption{Visualization of the $\mathbf{\alpha}$ table (0 is dark blue, 1 is
    yellow) for cognate detection using a unigram model. Left: A \emph{cognate}
    pair, Right: a \emph{non-cognate} pair }\label{fig:alphacognates}

\end{figure}
% % % % % % % % % % % % % % % % % % % % % % % % % % % % % % % % % % % % % % %

% % % % % % % % % % % % % % % % % % % % % % % % % % % % % % % % % % % % % % %
\begin{table}

		\centering\scalebox{1.0}{\centering\footnotesize
\begin{tabular}{lc}
\toprule
    Loss functions & F$_1$ \\ \midrule

    Complete loss                                    & \R{97.1}{0.6} \\

    --- binary XENT for $\alpha_{m,n}$               & \R{96.1}{0.3} \\

    --- expectation-maximization (Alg.~\ref{alg:em}) & \R{96.3}{0.7} \\

\bottomrule

\end{tabular}
}

    \caption{Ablation study for loss function on Cognate classification with a
    model with RNN contextualizer.}\label{tab:cognatesAblation}

\end{table}
% % % % % % % % % % % % % % % % % % % % % % % % % % % % % % % % % % % % % % %

% ----------------------------------------------------------------------------
%\subsection{Cognate Detection}
\paragraph{Cognate Detection.}
% ----------------------------------------------------------------------------
%
The results of cognate detection are presented in Table~\ref{tab:cognates}
(learning curves are in Figure~\ref{fig:curves} in Appendix). In cognate
detection, our model significantly outperforms both the statistical baseline
and the STANCE model. The F$_1$-score achieved by the unigram model is worse
than the Transformer classifier by a large margin. Local representation
contextualization with CNN reaches similar performance as the black-box
Transformer classifier while retaining a similar strong interpretability to the
static embeddings. Models with RNN encoders outperform the baseline classifier,
whereas the Transformer encoder yields slightly worse results. Detecting
cognates seems to be more difficult in Austro-Asiatic languages than in
Indo-European languages. The training usually converges before finishing a
single epoch of the training data.
\enote{AF}{Would it be interesting to present Figure 2 for the CNN and/or the
strongly contextual model? Or just say how they look in the text?}
An example of how the $\mathbf{\alpha}$ captures the prefix-pair probabilities
is shown in Figure~\ref{fig:alphacognates}.
The interpretability loss only has a negligible (although mostly slightly
negative) influence on the accuracy, within the variance of training runs. The
ablation study on loss functions (Table~\ref{tab:cognatesAblation}) shows that
the binary cross-entropy plays a more important role. The EM loss alone works
remarkably well given that it was trained on positive examples only.

% % % % % % % % % % % % % % % % % % % % % % % % % % % % % % % % % % % % % % %
\begin{table*}
    %\newcommand{\R}[2]{#1 \scalebox{0.8}}
    %\resizebox{\textwidth}{!}{
    \centering\scalebox{0.775}{\centering\footnotesize
\begin{tabular}{ll c ccccc ccccccc}
    \toprule
    \multicolumn{2}{l}{\multirow{3}{*}{Method}} &
    \multirow{3}{*}{\rotatebox{90}{\# Param.~}} &
        \multicolumn{5}{c}{Arabic $\rightarrow$ English} &
        \multicolumn{7}{c}{CMUDict}
        \\
        \cmidrule(lr){4-8} \cmidrule(lr){9-15}

    \multicolumn{2}{l}{}       &  &
        \multicolumn{2}{c}{Plain} &
        \multicolumn{2}{c}{+ Interpret.\ loss} &
        \multirow{2}{*}{Time} &

        \multicolumn{3}{c}{Plain} &
        \multicolumn{3}{c}{+ Interpret.\ loss} &
        \multirow{2}{*}{Time} \\
        \cmidrule(lr){4-5} \cmidrule(lr){6-7} \cmidrule(lr){9-11} \cmidrule(lr){12-14}

    \multicolumn{2}{l}{}       &
        & CER & WER & CER & WER &
        & CER & WER & Align. & CER & WER & Align. \\ \midrule

    \multicolumn{2}{l}{RNN Seq2seq} & 3.3M
            & \R{22.0}{0.2} & \R{75.8}{0.6} & --- & --- & 12m
            & \R{ 5.8}{0.1} & \R{23.6}{0.9} & \A{24.5} & --- & --- & --- & 1.8h
            \\
    \multicolumn{2}{l}{Transformer} & 3.1M
            & \R{22.9}{0.2} & \R{78.5}{0.4} & --- & --- &  11m
            & \R{ 6.5}{0.1} & \R{26.6}{0.3} & \A{33.2} & --- & --- & --- & 1.1h
            \\
    \midrule

    \multirow{4}{*}{\rotatebox{90}{ours}}
    & unigram          & 0.7M
            & \R{31.7}{1.8} & \R{85.2}{0.9} & \R{31.2}{1.4} & \R{85.0}{0.5} & 36m
            & \R{20.9}{0.3} & \R{67.5}{1.0} & \A{55.7} & \R{20.6}{0.3} & \R{66.3}{0.2} & \A{59.5} & 2.4h
               \\
    & CNN (3-gram)     & 1.1M
            & \R{24.6}{0.6} & \R{80.5}{0.3} & \R{24.5}{0.9} & \R{80.1}{0.9} & 41m
            & \R{12.8}{1.0} & \R{48.4}{3.1} & \A{35.4} & \R{12.8}{0.2} & \R{48.4}{0.6} & \A{38.1} & 2.5h
               \\
    \cmidrule{2-15}
    & Deep CNN & 3.0M
            & \R{24.4}{0.5} & \R{80.0}{0.7} & \R{23.8}{0.3} & \R{79.3}{0.1} & 52m
            & \R{10.8}{0.5} & \R{41.4}{1.9} & \A{23.3} & \R{10.8}{0.5} & \R{42.1}{1.6} & \A{28.8} & 2.5h
            \\
    & RNN              & 2.9M
            & \R{24.1}{0.2} & \R{77.0}{2.0} & \R{22.0}{0.3} & \R{77.4}{0.8} & 60m
            & \R{ 7.8}{0.3} & \R{31.9}{1.3} & \A{44.7} & \R{7.3}{0.4} & \R{33.3}{1.5} & \A{48.9} & 2.3h
            \\
    %& \quad+ attention & 3.3M
    %        & \bf\R{22.8}{0.3} & \bf\R{76.9}{0.6} & \R{23.3}{0.5} & \R{77.6}{1.2} & 2.4h
    %        & \R{ 9.0}{0.4} & \R{36.6}{1.6} & \A{1.8} & \bf\R{ 8.9}{0.6} & \bf\R{36.4}{2.0} & \A{8.2} & 5.5h
    %        \\
    & Transformer      & 3.2M
            & \R{24.3}{0.9} & \R{79.0}{0.7} & \R{23.9}{1.6} & \R{78.6}{1.3} & 1.2h
            & \R{10.7}{1.0} & \R{41.8}{3.1} & \A{33.3} & \R{10.2}{1.1} & \R{43.6}{3.2} & \A{37.9} & 2.3h
            \\
    %& \quad+ attention & 3.7M
    %        & \R{23.0}{0.5} & \R{77.0}{0.8} & \R{23.3}{0.3} & \R{78.0}{0.8} & 2.5h
    %        & \R{10.7}{0.3} & \R{42.3}{1.0} & \A{1.7} &\R{10.7}{0.3} & \R{42.9}{0.5} & \A{1.7} & 6.8h
    %        \\

    \bottomrule
\end{tabular}
}

    \caption{Model error rates for Arabic-to-English transliteration and
    English G2P generation and respective training times. For the second data
    set, we also report the alignment F$_1$ scores (Align.). Our best models
    are in bold.  The error rates on the validation data are in
    Table~\ref{tab:transliteration_valid} in the
    Appendix.}\label{tab:transliteration}

\end{table*}
% % % % % % % % % % % % % % % % % % % % % % % % % % % % % % % % % % % % % % %

% % % % % % % % % % % % % % % % % % % % % % % % % % % % % % % % % % % % % % %
\begin{figure}

    \includegraphics{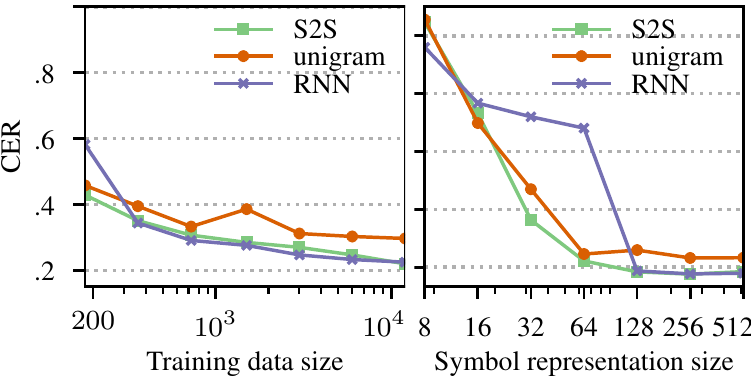}

    \caption{Character Error Rate for Arabic transliteration into English for
      %models trained training sets of
      \emph{various training data sizes} (left)
    and \emph{various representation sizes} (right).}\label{fig:datasizes}

\end{figure}
% % % % % % % % % % % % % % % % % % % % % % % % % % % % % % % % % % % % % % %

% ----------------------------------------------------------------------------
%\subsection{Transliteration and G2P Conversion}
\paragraph{Transliteration and G2P Conversion.}
% ----------------------------------------------------------------------------
%
The results for the two transduction tasks are presented in
Table~\ref{tab:transliteration} (learning curves are in Figure~\ref{fig:curves}
in Appendix). Our transliteration baseline slightly outperforms the baseline
presented with the dataset \citep[22.4\% CER, 77.1\% WER]{rosca2016sequence}.
Our baselines for the G2P conversion perform slightly worse than the best
models by \citet{yolchuyeva2019transformer},
%that reaches
which had
5.4\% CER and 22.1\%
WER with a twice as large model, and 6.5\% CER and 23.9\% WER
%for
with
a similarly
sized one.

The transliteration of Arabic appears to be a simpler problem than G2P
conversion. The performance matches S2S,
%reaches faster training times
has fast training times,
and
%the
there is a
smaller gap between the error rates of the context-free and contextualized
models.

The training time of our transduction models is 2--3$\times$ higher than with
the baseline S2S models because the baseline models use builtin PyTorch
functions, whereas our model is implemented using loops using
TorchScript\footnote{\url{https://pytorch.org/docs/stable/jit.html}} (15\%
faster than plain Python). The performance under low data conditions and with
small model capacity is in Figure~\ref{fig:datasizes}.

% % % % % % % % % % % % % % % % % % % % % % % % % % % % % % % % % % % % % % %
\begin{table}

    \centering\scalebox{1.0}{\centering\footnotesize

\begin{tabular}{lcc}
\toprule
    Loss functions & CER & WER \\ \midrule

    Complete loss                    & \R{22.5}{0.3} & \R{77.4}{0.8} \\
     --- expectation maximization    & \R{68.2}{7.4} & \R{93.5}{1.0} \\
     --- next symbol NLL             & \R{27.2}{1.4} & \R{81.1}{2.2} \\
     --- $\alpha_{m,n}$ maximization & \R{23.5}{1.3} & \R{79.2}{2.5} \\

\bottomrule
\end{tabular}
}

    \caption{Ablation study for loss function on Arabic-to-English
    transliteration using RNN and the underlying
    representation.}\label{tab:generationAblation}

\end{table}
% % % % % % % % % % % % % % % % % % % % % % % % % % % % % % % % % % % % % % %

Models that use static symbol embeddings as the input perform worse than the
black-box S2S models in both tasks. Local contextualization with CNN improves
the performance over static symbol embeddings. Using the fully contextualized
input representation narrows the performance gap between S2S models and neural
string edit distance models at the expense of decreased interpretability
because all input states can, in theory, contain information about the entire
input sequence.
%This can be seen on the ability to perserve source-target
%alignment, which is the highest 
The ability to preserve source-target alignment is highest
when the input is represented by embeddings only.
RNN models not only have the best accuracy, but also capture quite well
the source-target alignment. We hypothesize that RNNs work
%particularly
well
because
of their inductive bias towards sequence processing, which might be hard
to learn
from
position embeddings given the task dataset sizes.
%from the task.

Including the interpretability loss usually slightly improves the accuracy and
improves the alignment between the source and target strings. It manifests
both qualitatively (Table~\ref{tab:interpret}) and quantitatively in the
increased alignment accuracy.

Compared to S2S models, beam search decoding leads to much higher accuracy
gains, with beam search 5 reaching around 2$\times$ error reduction compared to
greedy decoding. For all input representations except the static embeddings,
length normalization does not improve decoding. Unlike machine translation
models,
%the accuracy does to
accuracy
%does not
doesn't
degrade with increasing beam size. See
Figure~\ref{fig:beams} in
%the
Appendix.
%for details.

The ablation study on loss functions (Table~\ref{tab:generationAblation}) shows
that all loss functions contribute to the final accuracy. The EM loss
%plays the
is
most important,
%role,
direct optimization of the likelihood is
%the second most important.
second.
%
% The last sentence is here becase the reviewers criticize that we do not try
% optimizing the likelihood.

% ============================================================================
\section{Related Work}
% ============================================================================

\paragraph{Weighted finite-state transducers.}%
\citet{rastogi-etal-2016-weighting} use a weighted-finite state transducer
(WFST) with neural scoring function to model sequence transduction. As in our
model, they back-propagate the error via a dynamic program. Our model is
stronger because, in the WFST, the output symbol generation only depends on the
contextualized source symbol embedding, disregarding the string generated so
far.

% % % % % % % % % % % % % % % % % % % % % % % % % % % % % % % % % % % % % % %
\begin{table*}

\newcommand{\graphemes}[1]{\multirow{2}{*}{\textcolor{Blue}{#1}}}
\newcommand{\phonemes}[1]{\multirow{2}{*}{\textcolor{Green}{#1}}}

    \centering
    %\resizebox{.99\columnwidth}{!}{%
    \footnotesize
    \begin{tabular}{llc}\toprule
        graphemes & phonemes & edit operations \\ \midrule

\graphemes{GOELLER} & \phonemes{G OW L ER}
& \subs{G}{G} \del{O} \del{E} \del{L} \subs{L}{OW} \ins{L} \del{E} \subs{R}{ER} \\
& & \subs{G}{G} \subs{O}{OW} \del{E} \del{L} \subs{L}{L} \del{E} \subs{R}{ER} \\
\midrule

\graphemes{VOGAN} & \phonemes{V OW G AH N}
& \subs{V}{V} \del{O} \subs{G}{OW} \ins{G} \ins{AH} \del{A} \subs{N}{N}  \\
& & \subs{V}{V} \ins{OW} \del{O} \subs{G}{G} \del{A} \subs{N}{N} \\
\midrule

\graphemes{FLAGSHIPS} & \phonemes{F L AE G SH IH P S}
& \subs{F}{F} \subs{L}{L} \del{A} \del{G} \subs{S}{AE} \ins{G} \del{H} \ins{SH} \del{I} \subs{P}{IH} \ins{P} \ins{S} \\
& & \subs{F}{F} \subs{L}{L} \ins{AE} \del{A} \subs{G}{G} \del{S} \subs{H}{SH} \ins{IH} \del{I} \subs{P}{P} \subs{S}{S}  \\
\midrule

\graphemes{ENDLER} & \phonemes{EH N D L ER}
& \ins{EH} \del{E} \subs{N}{N} \subs{D}{D} \subs{L}{L} \del{E} \subs{R}{ER}  \\
& & \subs{E}{EH} \subs{N}{N} \subs{D}{D} \subs{L}{L} \del{E} \subs{R}{ER}  \\
\midrule

\graphemes{SWOOPED} & \phonemes{S W UW P T}
& \subs{S}{S} \subs{W}{W} \ins{UW} \del{O} \del{O} \subs{P}{P} \del{E} \subs{D}{T} \\
& & \subs{S}{S} \subs{W}{W} \del{O} \subs{O}{UW} \subs{P}{P} \del{E} \subs{D}{T} \\

    \bottomrule
    \end{tabular}%}

    \caption{Edit operations predicted by
      %the
      RNN-based model for grapheme
        (blue) to phoneme (green) conversion with and without the
        interpretability loss (when provided ground-truth target). Green boxes
        %denote
are
insertions, blue boxes deletions,
%and
yellow boxes
        substitutions.}\label{tab:interpret}

\end{table*}
% % % % % % % % % % % % % % % % % % % % % % % % % % % % % % % % % % % % % % %

\citet{lin-etal-2019-neural} extend the model by including contextualized
target string representation and edit operation history. This makes their model
%stronger
more powerful
than
%our model.
ours,
%On the other hand,
but
the loss function
%can no longer be
cannot be exactly
computed
%exactly
by dynamic programming and requires sampling possible
operation sequences.

\paragraph{Segment to Segment Neural Transduction.} \citet{yu-etal-2016-online}
use two operation algorithm (shift and emit) for string transduction.  Unlike
our model directly, it models independently the operation type and target
symbols and lacks the concept of symbol substitution.

\paragraph{Neural sequence matching.}%
Several neural sequence-matching methods utilize a scoring function similar to
symbol-pair representation. \citet{cuturi2017soft} propose integrating
alignment between two sequences into a loss function that eventually leads to
finding alignment between the sequences.
The STANCE model \citep{tam-etal-2019-optimal}, which we compare
%our
results
with, first computes the alignment as an optimal transfer problem between the
source and target representation. In the second step, they assign a score using
a convolutional neural network applied to a soft-alignment matrix. We showed
that our model reaches better accuracy with the same input representation.
Similar to our model, these approaches provide interpretability via alignment.
They allow many-to-many alignments, but cannot enforce a monotonic sequence of
operations unlike WFSTs and our model.

\paragraph{Learnable edit distance.}%
\citet{mccallum2005conditional} used trainable edit distance in combination
with
%conditional random fields
CRFs
for string matching. Recently,
\citet{riley2020unsuupervised} integrated the statistical learnable edit
distance within a pipeline for unsupervised bilingual lexicon induction. As far
as we know, our work is the first
%attempt to use
using
neural networks directly in
%the
dynamic programming for edit distance.

\paragraph{Edit distance in deep learning.}%
LaserTagger \citep{malmi-etal-2019-encode} and EditNTS
\citep{dong-etal-2019-editnts} formulate sequence generation as tagging of the
source text with edit operations. They use standard edit distance to
pre-process the data (so, unlike our model cannot work with different
alphabets) and then learn to predict the edit operations.
Levenshtein Transformer \citep{gu2019levenshtein} is a partially
non-autoregressive S2S model generating the output iteratively via insert and
delete operations. It delivers a good trade-off of decoding speed and
translation quality, but is not interpretable.
%
% TODO mention EDITOR in camera ready

\paragraph{Dynamic programming in deep learning.}%
Combining dynamic programming and neural-network-based estimators is a common
technique, especially in sequence modeling. Connectionist Temporal
Classification (CTC; \citealp{graves2006connectionist}) uses the
forward-backward algorithm to estimate the loss of assigning labels to a
sequence with implicit alignment. The loss function of a linear-chain
conditional random field propagated into a neural network \citep{do2010neural}
became the state-of-the-art for tasks like named entity recognition
\citep{lample-etal-2016-neural}.
Loss functions based on dynamic programming are also used in
non-autoregressive neural machine translation
\citep{libovicky-helcl-2018-end,saharia2020non}.

\paragraph{Cognate detection.}%
Due to the limited amount of annotated data, cognate detection is usually
approached using unsupervised methods. Strings are compared using measures such
as pointwise mutual information \citep{jager2014phylogenetic} or
%LextStat
LexStat
similarity \citep{list-2012-lexstat}, which are used as an input to a
distance-based clustering algorithm \citep{list-etal-2016-using}.
\citet{jager-etal-2017-using} used a supervised SVM classifier trained on one
language family using features that were previously used for clustering and
applied the classifier to other language families.

\paragraph{Transliteration.}%
Standard S2S models
\citep{bahdanau2015neural,gehring2017convolutional,vaswani2017attention} or
CTC-based sequence-labeling \citep{graves2006connectionist} are the state of
the art for both transliteration \citep{rosca2016sequence,kundu-etal-2018-deep}
and G2P conversion
\citep{yao2015sequence,peters-etal-2017-massively,yolchuyeva2019transformer}.

% ============================================================================
\section{Conclusions}
% ============================================================================

We introduced neural string edit distance, a neural model of string
transduction based on
%the
string edit distance.
%\citep{levenshtein1966binary,ristad1998learning}.
Our novel formulation of
neural string edit distance critically depends on a differentiable loss. When
used with context-free representations, it offers a direct interpretability via
insert, delete and substitute operations, unlike widely used
S2S models. Using input representations with differing
amounts of contextualization, we can trade off interpretability for better
performance.
Our experimental results on cognate detection, Arabic-to-English
transliteration and grapheme-to-phoneme conversion show that with
contextualized input representations, the proposed model is able to match the
performance of standard black-box models.
We hope that
our approach will help motivate more work on this type of interpretable model
and that our framework will be useful in such future work.

\section*{Acknowledgments}

The work at LMU Munich was supported by was supported by the European Research
Council (ERC) under the European Union’s Horizon 2020 research and innovation
programme (No.~640550) and by the German Research Foundation (DFG; grant FR
2829/4-1). The work at CUNI was supported by the European Commission via its
Horizon 2020 research and innovation programme (No.~870930).

\bibliography{anthology,references}
\bibliographystyle{acl_natbib}

%\clearpage
\appendix

\newwrite\outputstream
\immediate\openout\outputstream=appendix.tmp
\immediate\write\outputstream{\thepage}
\immediate\closeout\outputstream

%\onecolumn
%\vspace{-0.6cm}\begin{multicols}{2}

% ----------------------------------------------------------------------------
\section{Inference algorithm}\label{ap:alg}
% ----------------------------------------------------------------------------

\enote{AF}{It might be a good idea to say what is in the appendix in the main
paper if you can (I think you mention one section currently, and the others
are not mentioned)}

Algorithm~\ref{alg:alpha} is a procedural implementation of
Equation~\ref{eq:alpha}. In the Viterbi decoding used for obtaining the
alignment, the summation on line 6, 8 and 10 is replaced by taking the
maximum.

% % % % % % % % % % % % % % % % % % % % % % % % % % % % % % % % % % % % % % %
\begin{algorithm}[h]
\caption{Forward evaluation}\label{alg:alpha}
\small
\begin{algorithmic}[1]
    \State $\mathbf{\alpha} \in \mathbb{R}^{n\times m} \gets \mathbf{0}$
    \State $\alpha_{0,0} \gets 1$
    \For{$i = 1 \ldots n$}
        \For{$j = 1 \ldots m$}
            \If{$j > 0$}
                \State $\alpha_{i,j} \pluseq \Pins(t_j|\mathbf{c}_{i,j-1}) \cdot \alpha_{i, j-1}$
            \EndIf
            \If{$i > 0$}
                \State $\alpha_{i,j} \pluseq \Pdel(s_i|\mathbf{c}_{i-1,j}) \cdot \alpha_{i-1, j}$
            \EndIf
            \If{$i > 0$ and $j > 0$}
                \State $\alpha_{i,j} \pluseq \Psubs(s_i\subsarrow t_j | \mathbf{c}_{i-1,j-1}) \cdot \alpha_{i-1,j-1}$
            \EndIf
        \EndFor
    \EndFor

\end{algorithmic}
\end{algorithm}
% % % % % % % % % % % % % % % % % % % % % % % % % % % % % % % % % % % % % % %

% ----------------------------------------------------------------------------
\section{Motivation for design choices in the string-matching
    model}\label{ap:motivation}
% ----------------------------------------------------------------------------

Let us assume a toy example transliteration. The source alphabet is
$\{\mathtt{A}, \mathtt{B}, \mathtt{C}\}$, the target alphabet is
$\{\mathtt{a}, \mathtt{b}, \mathtt{c}\}$, the transcription rules are:
\begin{enumerate}
    \item If $\mathtt{B}$ is at the beginning of the string, delete it.
    \item Multiple $\mathtt{A}$s rewrite to a single $\mathtt{a}$.
    \item Rewrite $\mathtt{B}$ to $\mathtt{b}$ and $\mathtt{C}$ to $\mathtt{c}$.
\end{enumerate}

The statistical learnable edit distance would not be capable of properly
learning rules 1 and 2 because it would not know that $\mathtt{B}$ was at the
beginning of the string and
%how
if an occurrence of $\mathtt{A}$ is the first
%one.
$\mathtt{A}$.
This problem gets resolved by introducing a contextualized
representation of the input.

The original statistical EM algorithm only needs positive examples to learn the
operation distribution. For instance, rewriting $\mathtt{B}$ to $\mathtt{c}$
will end up as improbable due to the inherent limitation of a single sharing
static probability table. Using a single table regardless of the context means
that if some operations become more probable, the
%other
others
must become less
probable. A neural network does not have such limitations. A neural model can
in theory find solutions that maximize the probability of the training data,
however, do not correspond to the original set of rules by finding a highly
probable sequence of operations for any string pair. For instance, it can learn
to count the positions in the string:
\begin{itemize}
    \item[1$'$.] Whatever symbols at the same position $i$ ($s_i$ and $t_i$)
        are, substitute $s_i$ with $t_j$ with the probability of 1.
    \item[2$'$.] If $i < j$, assign probability of 1 to deleting $s_i$.
    \item[3$'$.] If $i > j$, assign probability of 1 to inserting $t_j$.
\end{itemize}

For this reason, we introduce the binary cross-entropy as an additional loss
function.
%,
%that
This
should steer the model away from degenerate solutions assigning
a high probability score to any input string pair.

%As
But
our ablation study in Table~\ref{tab:cognatesAblation} showed that even
without the binary cross-entropy loss, the model converges to a good
non-degenerate solution.

This thought experiment shows keeping the full table of possible model outcomes
is no longer crucial for the modeling strength. Let us assume that the output
distribution of the neural model contains all possible edit operations as they
are in the static probability tables of the statistical model.
The model can learn to rely on the position information only and select the
correct symbols in the output probability distribution ignoring the actual
content of the symbols, using their embeddings as a key to identify the correct
item from the output distribution. The model can thus learn to ignore the
function the full probability table had in the statistical model.
Also, given the inputs, it is always clear what the plausible operations are, it
is easy for the model not to assign any probability to the implausible
%operation
operations
(unlike the statistical model).

These thoughts lead us to
%a
the
conclusion that there is no need to keep the full
output distribution and we only can use four target classes: one for insertion,
one for deletion, one for substitution, and one special class that would get
the part of probability mass that would be assigned to implausible operations
in the statistical model. We call the last one the \emph{non-match} option.

% ----------------------------------------------------------------------------
\section{Model Hyperparameters}\label{ap:hyper}
% ----------------------------------------------------------------------------

Following \citet{gehring2017convolutional}, the CNN uses gated linear units as
non-linearity \citep{dauping2017language}, layer normalization
\citep{ba2015layer} and residual connections \citep{he2016deep}. The symbol
embeddings are summed with learnable position embeddings before the
convolution.

The RNN uses gated recurrent units \citep{cho-etal-2014-learning} and follows
the scheme of \citet{chen-etal-2018-best}, which includes residual connections
\citep{he2016deep}, layer normalization \citep{ba2015layer}, and multi-headed
scaled dot-product attention \citep{vaswani2017attention}.

The Transformers follow the architecture decisions of BERT
\citep{devlin-etal-2019-bert} as implemented in the Transformers library
\citep{wolf2019huddingface}.

All hyperparameters are set manually based on preliminary experiments.  For all
experiments, we use embedding size of 256. The CNN encoder uses a single
layer with kernel size 3 and ReLU non-linearity. For both the RNN and
Transformer models, we use 2 layers with 256 hidden units. The Transformer
uses 4 attention heads of dimension 64 in the self-attention.  The same
configuration is used for the encoder-decoder attention for both RNN and
Transformer. We use the same hyperparameters also for the baselines.

We include all main loss functions with weight 1.0, i.e., for string-pair
matching: the EM loss, non-matching negative log-likelihood and binary
cross-entropy; for string transduction: the EM loss and next symbol negative
log-likelihood. We test each model with and without the interpretability loss,
which is included with weight 0.1.

We optimize the models using the Adam optimizer \citep{kingma2015adam} with an
initial learning rate of $10^{-4}$, and batch size of 512. We validate the
models every 50 training steps. We decrease the learning rate by a factor of
$0.7$ if the validation performance does not increase in two consecutive
validations. We stop the training after the learning rate decreases 10 times.

% ----------------------------------------------------------------------------
\section{Notes on Reproducibility}\label{ap:reproducibility}
% ----------------------------------------------------------------------------

The training times were measured on machines with GeForce GTX 1080 Ti GPUs and
with Intel Xeon E5--2630v4 CPUs (2.20GHz). We report average wall time of
training including data preprocessing, validation and testing. The measured
time might be influenced by other processes running on the machines.

Validation scores are provided in Tables~\ref{tab:cognates_valid}
and~\ref{tab:transliteration_valid}.

%\end{multicols}

% % % % % % % % % % % % % % % % % % % % % % % % % % % % % % % % % % % % % % %
\begin{table*}[h]
    \centering\footnotesize
\begin{tabular}{lllcc cc}
\toprule
\multicolumn{2}{l}{\multirow{2}{*}{Method}} &
    \multicolumn{2}{c}{Indo-European} &
    \multicolumn{2}{c}{Austro-Asiatic} \\
    \cmidrule(lr){3-4} \cmidrule(lr){5-6}

\multicolumn{2}{l}{}       &
    Base & + Int.\ loss &
    Base & + Int.\ loss \\ \midrule

\multicolumn{2}{l}{Transformer \tt[CLS]} &
    \R{91.4}{2.8} & ---   &
    \R{78.8}{0.8} & ---   \\ \midrule

\multirow{3}{*}{\rotatebox{90}{\scriptsize STANCE~~}}
& unigram     &
	\R{46.5}{4.7} & --- &
    \R{16.5}{0.4} & --- \\ \cmidrule{2-6}
& RNN         &
    \R{80.4}{1.6} & --- &
    \R{16.5}{0.1} & ---  \\
& Transformer &
    \R{76.8}{1.3} & --- &
    \R{17.2}{0.2} & ---  \\ \midrule

\multirow{4}{*}{\rotatebox{90}{ours}}
& unigram     &
    \R{81.2}{1.0} & \R{82.0}{0.5}  &
    \R{52.6}{0.8} & \R{53.9}{0.6}  \\
& CNN (3-gram)     &
    \R{95.2}{0.6} & \R{94.9}{0.7}  &
    \R{78.9}{0.8} & \R{78.1}{1.7}  \\ \cmidrule{2-6}
& RNN         &
    \R{97.2}{0.2} & \R{88.8}{1.1}  &
    \R{82.8}{0.6} & \R{83.1}{0.7}  \\
& Transformer &
    \R{88.8}{1.6} & \R{88.7}{1.1}  &
    \R{71.5}{1.1} & \R{71.5}{1.1}  \\
\bottomrule
\end{tabular}

    \caption{F$_1$-score for cognate detection on the \emph{validation}
    data.}\label{tab:cognates_valid}

\end{table*}
% % % % % % % % % % % % % % % % % % % % % % % % % % % % % % % % % % % % % % %

% % % % % % % % % % % % % % % % % % % % % % % % % % % % % % % % % % % % % % %
\begin{table*}[h]
    \centering\footnotesize
\begin{tabular}{ll cccc cccc}
\toprule
\multicolumn{2}{l}{\multirow{2}{*}{Method}} &
	\multicolumn{4}{c}{Arabic $\rightarrow$ English} &
	\multicolumn{4}{c}{CMUDict}
	\\
            \cmidrule(lr){3-6} \cmidrule(lr){7-10}

\multicolumn{2}{l}{}       &
	\multicolumn{2}{c}{Base} &
	\multicolumn{2}{c}{+ Int.\ loss} &

	\multicolumn{2}{c}{Base} &
	\multicolumn{2}{c}{+ Int.\ loss}  \\
	\cmidrule(lr){3-4} \cmidrule(lr){5-6} \cmidrule(lr){7-8} \cmidrule(lr){9-10}

\multicolumn{2}{l}{}        & CER & WER & CER & WER & CER & WER & CER & WER \\ \midrule

% ---------------------------- |  CMUDict    |  Transliteration
\multicolumn{2}{l}{RNN Seq2seq}
		& \R{21.7}{0.1} & \R{75.0}{0.6} & --- & ---
		& \R{ 7.4}{0.0} & \R{31.5}{0.1} & --- & ---
		\\
\multicolumn{2}{l}{Transformer}
		& \R{22.8}{0.2} & \R{77.7}{0.6} & --- & ---
		& \R{ 7.8}{0.1} & \R{32.7}{0.3} & --- & ---
		\\
\midrule

% ---------------------------- |  CMUDict    |  Transliteration
\multirow{6}{*}{\rotatebox{90}{ours}}
& unigram
		& \R{28.4}{0.7} & \R{84.1}{0.8} & \R{28.3}{0.5} & \R{84.3}{0.7}
		& \R{21.2}{1.0} & \R{66.4}{1.9} & \R{21.5}{0.8} & \R{68.0}{2.1}
		\\
& CNN (3-gram)
		& \R{34.4}{1.1} & \R{86.5}{0.8} & \R{32.2}{1.1} & \R{86.5}{0.8}
		& \R{36.0}{5.7} & \R{80.9}{3.2} & \R{33.8}{3.5} & \R{79.0}{2.8}
		\\
\cmidrule{2-10}
& RNN
		& \R{42.4}{9.0} & \R{90.9}{5.4} & \R{45.2}{2.6} & \R{90.9}{1.8}
		& \R{59.1}{2.5} & \R{96.2}{0.7} & \R{43.6}{5.6} & \R{80.5}{5.6}
		\\
%& \quad+ attention
%		& \R{22.4}{0.7} & \R{76.7}{0.9} & \R{22.4}{0.5} & \R{76.9}{1.2}
%		& \R{ 9.2}{0.7} & \R{37.5}{1.8} & \R{ 9.3}{0.6} & \R{37.9}{1.7}
%		\\
& Transformer
		& \R{41.2}{9.1} & \R{91.7}{4.4} & \R{47.7}{3.6} & \R{92.5}{2.4}
		& \R{24.6}{4.3} & \R{73.8}{6.1} & \R{43.5}{3.6} & \R{84.9}{2.5}
		\\
%& \quad+ attention
%		& \R{22.9}{0.7} & \R{77.5}{1.7} & \R{78.3}{1.5} & \R{23.0}{0.8}
%		& \R{11.2}{0.4} & \R{44.2}{1.3} & \R{12.1}{1.2} & \R{45.6}{0.7}
%		\\
% ---------------------------- |  CMUDict    |  Transliteration

\bottomrule
\end{tabular}

    \caption{Model error-rates for Arabic-to-English transliteration and
    English G2P generation on \emph{validation}
    data.}\label{tab:transliteration_valid}.

\end{table*}
% % % % % % % % % % % % % % % % % % % % % % % % % % % % % % % % % % % % % % %

\begin{figure*}
    \centering

%    \begin{minipage}{.33\textwidth}
%        \centering
%    \small RNN Sequence-to-sequence
%
%    \input{plots/beam_s2s.pgf}
%
%    \end{minipage}\begin{minipage}{.33\textwidth}
%        \centering
%
%    \small Ours w/ static embeddings
%
%    \input{plots/beam_embeddings.pgf}
%
%    \end{minipage}\begin{minipage}{.33\textwidth}
%        \centering
%
%    \small Ours w/ shallow CNN
%
%    \input{plots/beam_cnn.pgf}
%
%    \end{minipage}
%
%    % SECOND LINE OF BEAM SEARCH GRAPHS
%    \begin{minipage}{.33\textwidth}
%        \centering
%    \small Ours w/ deep CNN
%
%    \input{plots/beam_cnndeep.pgf}
%
%    \end{minipage}\begin{minipage}{.33\textwidth}
%        \centering
%
%    \small Ours w/ RNN
%
%    \input{plots/beam_rnn.pgf}
%
%    \end{minipage}\begin{minipage}{.33\textwidth}
%        \centering
%
%    \small Ours w/ Transformer
%
%    \input{plots/beam_transformer.pgf}
%
%    \end{minipage}

    \includegraphics{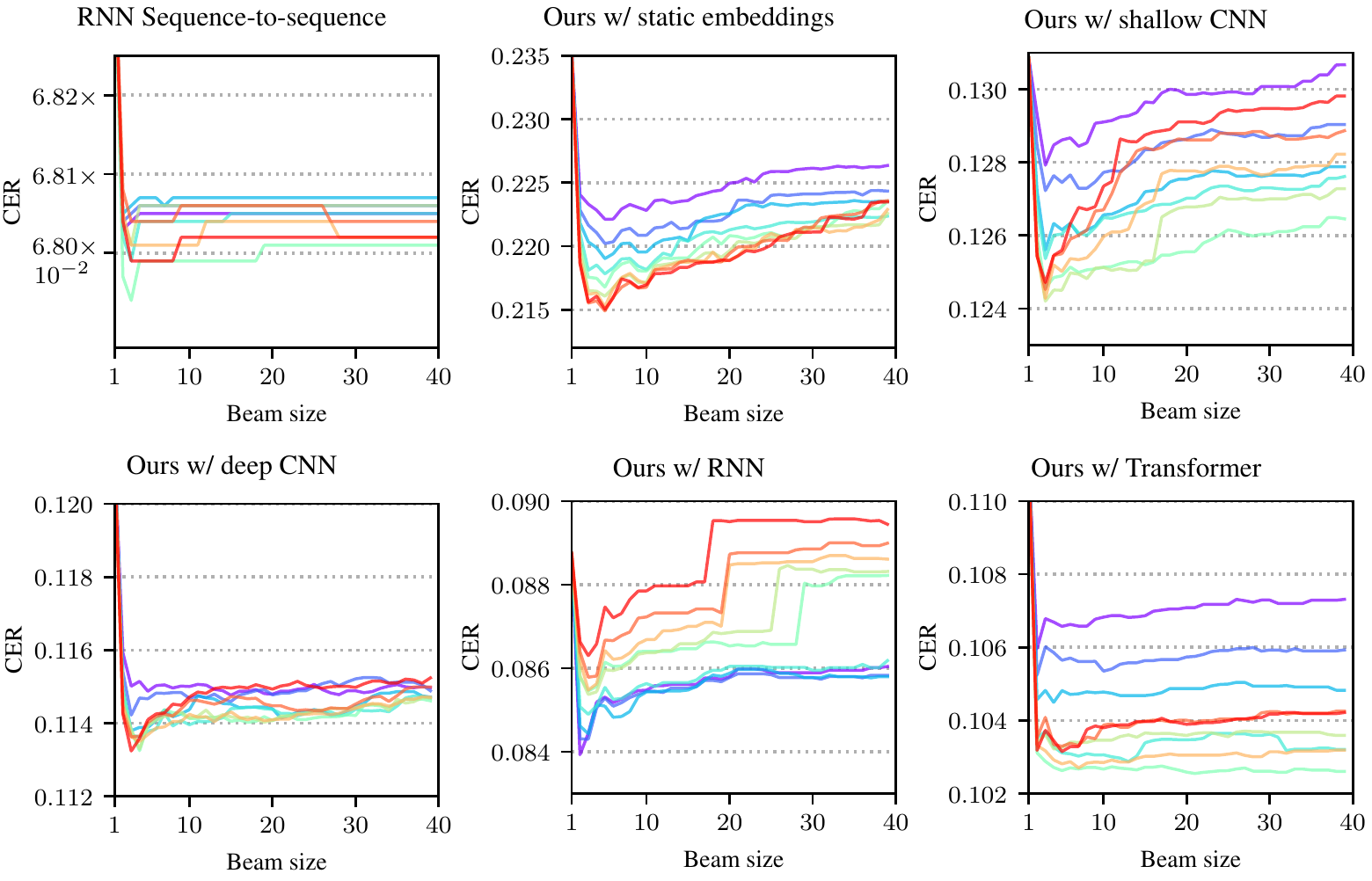}

%    % SECOND LINE OF BEAM SEARCH GRAPHS
%    \begin{minipage}{.33\textwidth}
%        \centering
%    \small Ours w/ deep CNN + attention
%
%    \input{plots/beam_cnndeep_att.pgf}
%
%    \end{minipage}\begin{minipage}{.33\textwidth}
%        \centering
%
%    \small Ours w/ RNN + attention
%
%    \input{plots/beam_rnn_att.pgf}
%
%    \end{minipage}\begin{minipage}{.33\textwidth}
%        \centering
%
%    \small Ours w/ Transformer + attention
%
%    \input{plots/beam_transformer_att.pgf}
%
%    \end{minipage}

    % THIS HOW THE COLOR CODES WERE GENERATED
%import matplotlib.pyplot as plt
%for val, alpha in zip([0.0, 0.2, 0.4, 0.6, 0.8, 1.0, 1.2, 1.4, 1.6], "ABCDEDGHI"):
%    r, g, b, _ = plt.cm.rainbow(val / 1.6)
%    print(f"\\definecolor{{lennorm{alpha}}}{{rgb}}{{{r:5f}, {g:5f}, {b:5f}}}")

    \definecolor{lennormA}{rgb}{0.500000, 0.000000, 1.000000}
    \definecolor{lennormB}{rgb}{0.249020, 0.384106, 0.980635}
    \definecolor{lennormC}{rgb}{0.001961, 0.709281, 0.923289}
    \definecolor{lennormD}{rgb}{0.245098, 0.920906, 0.833602}
    \definecolor{lennormE}{rgb}{0.503922, 0.999981, 0.704926}
    \definecolor{lennormF}{rgb}{0.754902, 0.920906, 0.552365}
    \definecolor{lennormG}{rgb}{0.998039, 0.709281, 0.384106}
    \definecolor{lennormH}{rgb}{1.000000, 0.384106, 0.195845}
    \definecolor{lennormI}{rgb}{1.000000, 0.000000, 0.000000}

    \newcommand{\colormark}[1]{\textcolor{#1}{$\blacksquare$}}

    \small Lenght normalization: \quad
    \colormark{lennormA} 0.0 \quad
    \colormark{lennormB} 0.2 \quad
    \colormark{lennormC} 0.4 \quad
    \colormark{lennormD} 0.6 \quad
    \colormark{lennormE} 0.8 \quad
    \colormark{lennormF} 1.0 \quad
    \colormark{lennormG} 1.2 \quad
    \colormark{lennormH} 1.4 \quad
    \colormark{lennormI} 1.6 \quad

    \caption{Effect of beam search on test data for grapheme-to-phoneme conversion.}\label{fig:beams}

\end{figure*}

% % % % % % % % % % % % % % % % % % % % % % % % % % % % % % % % % % % % % % %

\begin{figure*}
    \centering

    %\begin{minipage}{.5\textwidth}
    %    \centering
    %\small Cognate detection on IELEX

    %\input{plots/learning_curves_ielex_train.pgf}

    %\input{plots/learning_curves_ielex.pgf}

    %\end{minipage}\begin{minipage}{.5\textwidth}
    %    \centering

    %\small Grapheme-to-phoneme conversion

    %\input{plots/learning_curves_cmudict_train.pgf}

    %\input{plots/learning_curves_cmudict.pgf}

    %\end{minipage}
    \includegraphics{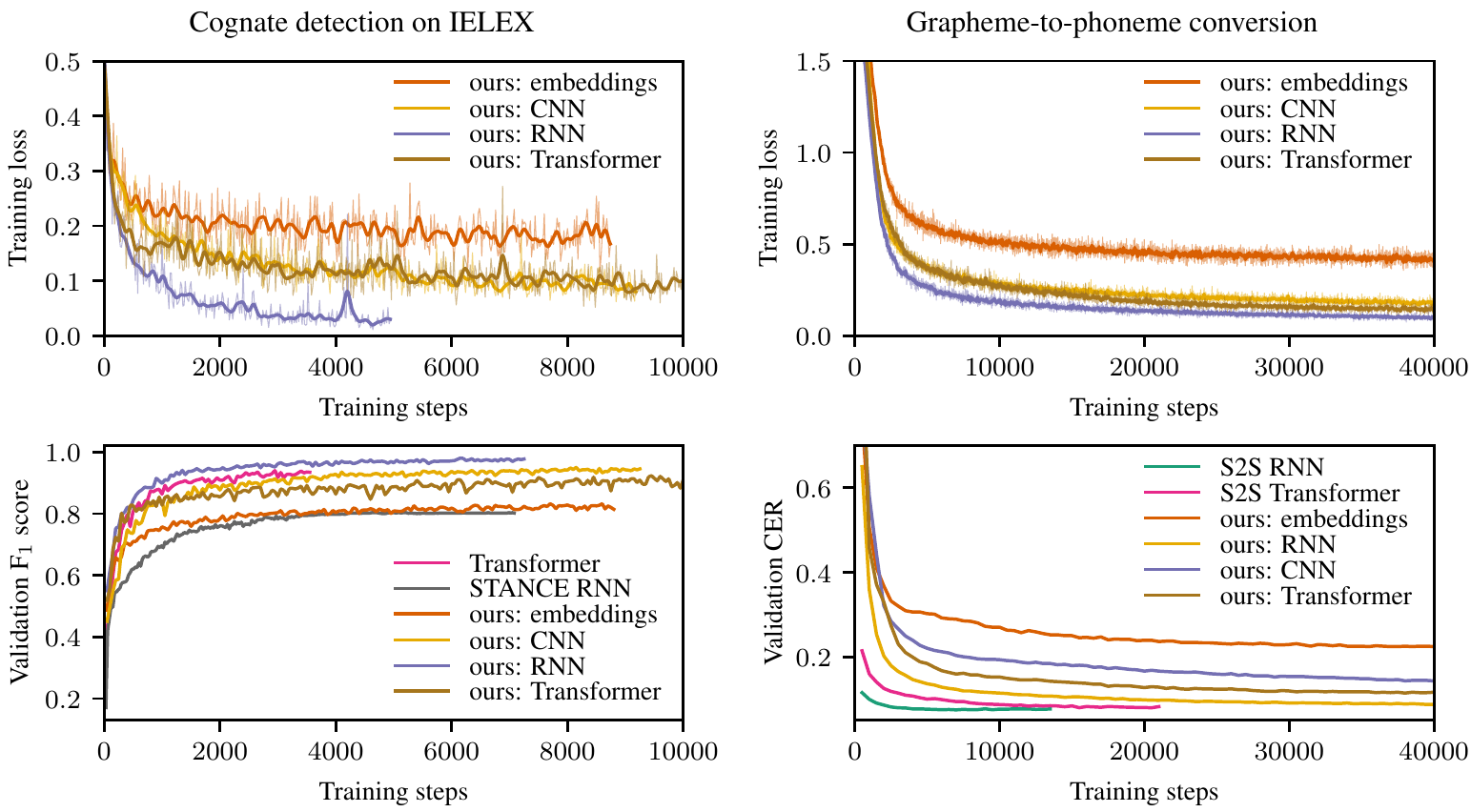}

    \caption{Learning curves for Cognate classification for Indo-European
    languages (left) and for grapheme-to-phoneme conversion (right).}\label{fig:curves}

\end{figure*}

\end{document}